# A Performance Investigation of Multimodal Multiobjective Optimization Algorithms in Solving Two Types of Real-World Problems

Zhiqiu Chen
*School of Computer Science*
*South China Normal University*
Guangzhou, China
1105122063@qq.com

Zong-Gan Chen
*School of Computer Science*
*South China Normal University*
Guangzhou, China
charleszg@qq.com

Yuncheng Jiang
*School of Computer Science*
*South China Normal University*
Guangzhou, China
ycjiang@scnu.edu.cn

Zhi-Hui Zhan
*College of Artificial Intelligence*
*Nankai University*
Tianjin, China
zhanapollo@163.com

*Abstract*—In recent years, multimodal multiobjective optimization algorithms (MMOAs) based on evolutionary computation have been widely studied. However, existing MMOAs are mainly tested on benchmark function sets such as the 2019 IEEE Congress on Evolutionary Computation test suite (CEC'2019), and their performance on real-world problems is neglected. In this paper, two types of real-world multimodal multiobjective optimization problems in feature selection and location selection respectively are formulated. Moreover, four real-world datasets of Guangzhou, China are constructed for location selection. An investigation is conducted to evaluate the performance of seven existing MMOAs in solving these two types of real-world problems. An analysis of the experimental results explores the characteristics of the tested MMOAs, providing insights for selecting suitable MMOAs in real-world applications.

## I. INTRODUCTION

Multimodal multiobjective optimization problems (MMOPs) are multiobjective optimization problems [1]–[3] where multiple Pareto sets (PSs) correspond to a Pareto front (PF). For instance, Fig. 1 depicts an MMOP with two decision variables and two objectives, where $PS_1$ and $PS_2$ are two PSs in the decision space, containing Pareto optimal solutions $X_1$ and $X_2$, respectively. $X_1$ and $X_2$ have identical objective values, corresponding to the same point on the PF in the objective space (i.e., $F(X_1) = F(X_2)$). A minimization MMOP can be defined as

$$\min F(X) = (f_1(X),...,f_M(X)),\ X = (x_1,...,x_D) \in \Omega \quad (1)$$

in which $X$ denotes a solution in $\Omega$ and $\Omega$ denotes the decision space. $F(X)$ denotes the objective value of $X$. The dimensions of decision space and the number of objectives are defined as $D$ and $M$, respectively.

This work was supported in part by the National Natural Science Foundations of China under Grant 62206100, in part by the Guangdong Basic and Applied Basic Research Foundation under Grant 2024A1515011708, in part by the Guangzhou Basic and Applied Basic Research Foundation under Grant 2023A04J0319, and in part by the Fundamental Research Funds for the Central Universities, Nankai University (078-63243159). *(Corresponding author: Zong-Gan Chen)*

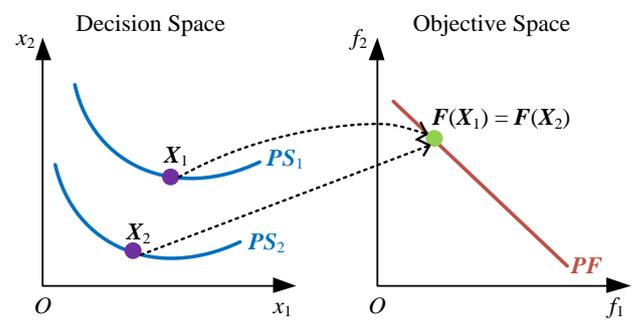

Fig. 1. Example of a minimization MMOP.

In MMOPs, a dominance relationship is used to compare the quality of two solutions. In detail, if a solution $X_1$ is not worse than another solution $X_2$ in any objective and is better than $X_2$ in at least one objective, $X_1$ is said to dominate $X_2$. Solutions in the decision space that are not dominated by any other solutions are called Pareto optimal solutions. The set of all Pareto optimal solutions forms the PSs. In addition, PF represents a set containing the objective values of all Pareto optimal solutions.

Numerous real-world problems are MMOPs [4], e.g., feature selection [5], [6] and credit card fraud detection [7], [8]. Evolutionary computation (EC) algorithms have been widely used to solve various optimization problems [9]–[12]. In recent years, extensive research is also conducted on multimodal multiobjective optimization algorithms (MMOAs) based on EC. To help MMOAs fully explore the decision space and identify as many Pareto optimal solutions as possible, methods for enhancing diversity can be categorized into three types (i.e., niching methods, crowding distance methods, and decomposition methods). Niching methods [13]–[15] aim to divide the entire population into multiple subpopulations to explore multiple PSs, and clustering is widely used to identify subpopulations [16]–[18]. Crowding distance methods [19]–[21] select individuals in sparser regions for evolution through the distance calculation between individuals. Decomposition methods [22]–[24] typically decompose the MMOP into multiple subproblems based on a set of uniformly distributed weight vectors. Due to the conflict between diversity and convergence, only focusing on enhancing diversity may lead to unsatisfactory solution accuracy. Thus, some MMOAs [25]–[28] have adopted methods to improve convergence and prevent the decline in convergence caused by an excessive focus on diversity.

The performance evaluation of existing MMOAs is mainly conducted on benchmark function sets of MMOPs, such as the 2019 IEEE Congress on Evolutionary Computation test suite (CEC'2019) [29], and existing MMOAs are rarely applied to real-world MMOPs. Therefore, we conduct experiments using four state-of-the-art and three classic MMOAs on two types of real-world problems: feature selection and location selection. The results reveal various characteristics of different algorithms, providing insights for selecting suitable MMOAs for specific MMOPs.

The remainder of the paper is structured as follows. Section II introduces the definitions of the two types of real-world problems. Section III describes the experimental setup in detail and analyzes the experimental results. Finally, Section IV concludes the paper.

## II. TWO TYPES OF REAL-WORLD MULTIMODAL MULTIOBJECTIVE OPTIMIZATION PROBLEMS

### A. Multimodal Multiobjective Feature Selection

Feature selection is important in data mining [30]. Selecting an appropriate feature subset from a large number of features can help achieve a promising classification error rate and improve the time efficiency. Previous studies on feature selection have primarily focused on its multiobjective characteristic, namely the two conflicting objectives of minimizing classification error rate and minimizing the number of selected features. However, in practice, there exist different feature subsets corresponding to the same objective values (referred to as equivalent feature subsets). Providing diverse feature subsets to the decision-maker can cater to the varying preferences and avoid high acquisition costs for some features.

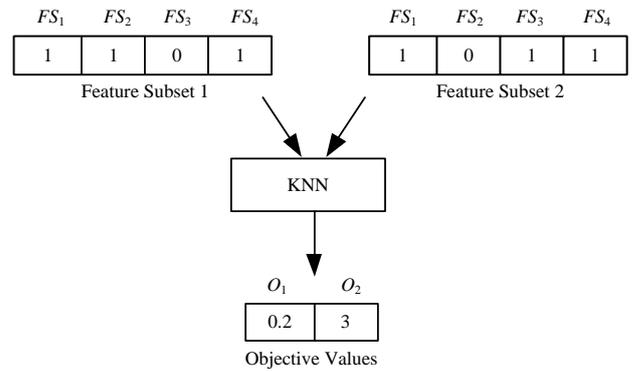

Fig. 2. Example of a feature selection problem.

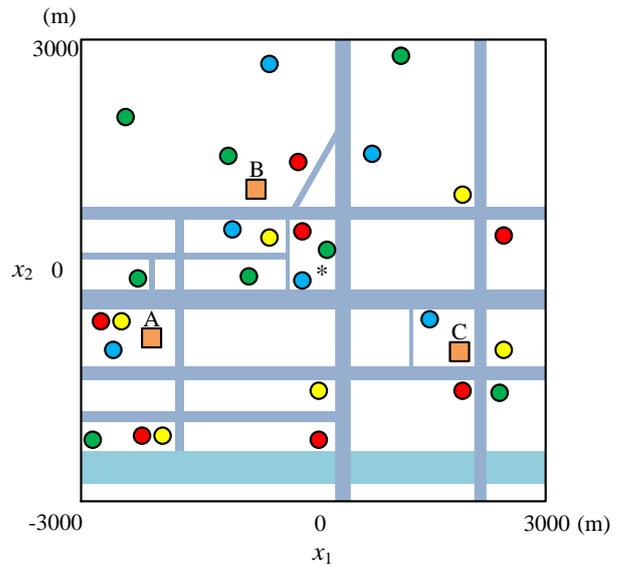

Fig. 3. Example of a location selection problem.

Fig. 2 shows an example of feature selection, in which there are two feature subsets. In detail, Feature Subset 1 selects three features (i.e., $FS_1$, $FS_2$, and $FS_4$), while Feature Subset 2 selects three features (i.e., $FS_1$, $FS_3$, and $FS_4$). These two feature subsets yield the same objective values $O_1$ (classification error rate) and $O_2$ (number of selected features) when calculated using the K-Nearest Neighbors (KNN) classifier. Therefore, the decision-maker has two feature selection options. If the cost of obtaining feature $FS_2$ is higher than that of feature $FS_3$, the decision-maker can choose Feature Subset 2 to achieve the same objective at a lower cost.

Thus, treating the feature selection problem as an MMOP and identifying equivalent feature subsets is imperative. Since classification difficulty is influenced by the number of samples, features, and classes, we select datasets with varying sizes of these factors to test the performance of seven algorithms in Section III.

### B. Multimodal Multiobjective Location Selection

In the location selection problem, we refer to the work of Ishibuchi *et al.* [31] and select four districts in Guangzhou, China (i.e., Tianhe District, Haizhu District, Yuexiu District, and Panyu District) to generate four real-world datasets. Specifically, in each district, we collect location data within a three-kilometer radius around a particular location. For example, in the dataset of Tianhe District, we suppose that a person works at South China Normal University (SCNU) in Tianhe District and needs to buy a house. Considering factors such as his children's education, living, and transportation convenience, the person wants to find a house within a three-kilometer radius of SCNU that is closest to the primary school, middle school, shopping center, and subway station.

Such a location selection problem is an MMOP. The four objectives to be minimized are the distances from the house to the nearest primary school, middle school, shopping center, and subway station. We use the distance intervals to measure objective values. In detail, 12 intervals (i.e., [0, 500), [500, 1000), [1000, 1500), [1500, 2000), [2000, 2500), [2500, 3000), [3000, 3500), [3500, 4000), [4000, 4500), [4500, 5000), [5000, 5500), and [5500, 6000]) are set and the corresponding objective values of these 12 intervals are 1, 2, 3, 4, 5, 6, 7, 8, 9, 10, 11, and 12, respectively. For example, if the distances from two houses to their corresponding nearest middle school are 1700 meters and 1800 meters, respectively, the two houses have the same objective value of 4 since the distances are both within the interval [1500, 2000).

Fig. 3 presents a simplified map of the area around SCNU. The asterisk represents SCNU, green circles represent primary schools, blue circles represent middle schools, yellow circles represent shopping centers, red circles represent subway stations, and orange squares represent the optimal housing locations obtained. Since the objective values are set to integers representing distance intervals, the four objective values for the three houses shown are equal. In practice, house A is located in the bustling city center of Tianhe District, where housing prices are very high; house B, although not in the city center, is in an excellent school district with a price slightly lower than house A; house C is more remote compared to houses A and B, and has the lowest price among the three. However, house C still meets the same objective values, providing good education, living, and transportation advantages. For those with limited budgets, purchasing house C is a more suitable choice. Solving the location selection problem as an MMOP can offer the decision-maker a variety of options that meet multiple objectives, which is highly practical.

## III. EXPERIMENTAL STUDIES

### A. Multimodal Multiobjective Optimization Algorithms in the Experiments

Seven existing MMOAs, including HREA [28], MMEA-WI [27], MMOEA/DC [16], CPDEA [25], TriMOEA-TA&R [26], MO_Ring_PSO_SCD [20], and Omni-optimizer [19], are tested to evaluate their performance on two types of real-world problems: feature selection and location selection. Details of these seven MMOAs are shown in Table I. Specifically, HREA employs a hierarchical ranking method to preserve global and local PFs based on the decision-maker's preferences. This method utilizes a local convergence judgment strategy and a strategy for retaining different levels of PFs to enhance diversity in both the decision space and the objective space. MMEA-WI assigns different weights to individuals based on their Euclidean distance in the decision space to calculate the crowding distance, thereby enhancing diversity. Additionally, it employs a convergence archive to preserve non-dominated solutions for participation in the generation of offspring, improving the algorithm's convergence. MMOEA/DC employs dual clustering in both the decision space and the objective space to preserve local PSs and enhance diversity in the objective space, respectively. After clustering in the decision space, it selects the non-dominated solutions within clusters and combines them with the remaining well-converged solutions in the objective space to form a temporary population. The temporary population is then clustered in the objective space and pruned until each cluster contains only one individual. CPDEA uses a convergence penalty density approach, where density is influenced by convergence status. First, CPDEA evaluates the local convergence quality of each individual based on the distances and dominance relationships among individuals. The local convergence quality is then used to derive transformed distances between individuals. The density calculated from the transformed distances serves as a criterion for individual selection, with CPDEA retaining only individuals with lower densities. TriMOEA-TA&R employs dual archive and recombination strategies. Decision variable analysis is used to obtain the convergence archive and the diversity archive. At the end of the evolution process, the Pareto optimal solutions are obtained by recombining the two archives. MO_Ring_PSO_SCD employs ring topology to create niches and uses a special crowding distance strategy to enhance diversity. Omni-optimizer employs a classic and straightforward diversity maintenance strategy by selecting non-dominated individuals with large crowding distances to generate new individuals.

TABLE I
LIST OF TESTED ALGORITHMS

| Algorithm | Year | Source |
|---|---|---|
| HREA | 2023 | *IEEE Transactions on Evolutionary Computation* |
| MMEA-WI | 2021 | *IEEE Transactions on Evolutionary Computation* |
| MMOEA/DC | 2021 | *IEEE Transactions on Evolutionary Computation* |
| CPDEA | 2020 | *IEEE Transactions on Evolutionary Computation* |
| TriMOEA-TA&R | 2019 | *IEEE Transactions on Evolutionary Computation* |
| MO_Ring_PSO_SCD | 2018 | *IEEE Transactions on Evolutionary Computation* |
| Omni-optimizer | 2005 | *International Conference on Evolutionary Multi-Criterion Optimization* |

TABLE II
DETAILS OF FEATURE SELECTION DATASETS

| Dataset | Name | Number of Samples | Number of Features | Number of Classes |
|---|---|---|---|---|
| $FS\text{-}D_1$ | Libras Movement | 360 | 90 | 15 |
| $FS\text{-}D_2$ | Multiple Features | 2000 | 240 | 10 |
| $FS\text{-}D_3$ | Isolet | 7797 | 617 | 26 |
| $FS\text{-}D_4$ | Darwin | 174 | 450 | 2 |
| $FS\text{-}D_5$ | Dry Bean Dataset | 13611 | 16 | 7 |
| $FS\text{-}D_6$ | Rice Cammeo Osmancik | 3810 | 7 | 2 |
| $FS\text{-}D_7$ | Glass | 214 | 9 | 7 |
| $FS\text{-}D_8$ | Zoo | 101 | 16 | 7 |

### B. Experimental Results on Multimodal Multiobjective Feature Selection

We conduct experiments on eight datasets from UCI [32]. These datasets vary in the number of samples, features, and classes, with specific details shown in Table II. Each dataset is randomly split, with 30% used as the test set, and 5-fold cross-validation KNN is used during the training process. To balance classification efficiency and accuracy, the k value for KNN is set to 5. We set the population size to 200, the maximum number of function evaluations to 20000, and conduct each experiment 21 times. The other parameters of each algorithm are set according to their corresponding references.

Hypervolume (HV) and the number of equivalent feature subsets are used as evaluation metrics. HV represents the area enclosed by the obtained PF and the reference point in the objective space. Since HV does not require the ideal PF as a reference, it is suitable for feature selection problems without an ideal PF. The objective values in the experiments are normalized, so the reference point for HV is set to (1,1). To make comparison results clearer, we present the results of 1/HV, where a smaller 1/HV value indicates better performance in the objective space. A higher number of equivalent feature subsets indicates better performance in the decision space. For example, if the same objective values correspond to $k$ different feature selection solutions, the number of equivalent feature subsets is recorded as $k$–1.

TABLE III
EXPERIMENTAL RESULTS OF AVERAGE 1/HV ON FEATURE SELECTION PROBLEMS

| Dataset | HREA | MMEA-WI | MMOEA/DC | CPDEA | TriMOEA-TA&R | MO_Ring_PSO_SCD | Omni-optimizer |
|---|---|---|---|---|---|---|---|
| $FS\text{-}D_1$ | 1.46(6) | **1.24(1)** | 1.28(4) | 1.25(2) | 1.26(3) | 1.40(5) | 2.03(7) |
| $FS\text{-}D_2$ | 1.31(5) | 1.09(2) | 1.20(3) | 1.26(4) | **1.06(1)** | 1.37(6) | 1.52(7) |
| $FS\text{-}D_3$ | 1.65(5) | 1.28(2) | **1.26(1)** | 1.60(3) | 1.93(7) | 1.63(4) | 1.88(6) |
| $FS\text{-}D_4$ | 1.62(3) | **1.24(1)** | 1.31(2) | 1.72(5) | 1.81(7) | 1.70(4) | 1.78(6) |
| $FS\text{-}D_5$ | **1.21(1)** | **1.21(1)** | **1.21(1)** | **1.21(1)** | **1.21(1)** | **1.21(1)** | **1.21(1)** |
| $FS\text{-}D_6$ | 1.16(3) | 1.15(2) | 1.15(2) | 1.15(2) | 1.15(2) | 1.15(2) | **1.14(1)** |
| $FS\text{-}D_7$ | 1.67(3) | 1.66(2) | 1.67(3) | 1.67(3) | 1.67(3) | **1.63(1)** | 1.66(2) |
| $FS\text{-}D_8$ | 1.15(2) | **1.14(1)** | **1.14(1)** | 1.15(2) | 1.15(2) | 1.15(2) | 1.15(2) |
| average ranking | 3.5(6) | **1.5(1)** | 2.125(2) | 2.75(3) | 3.25(5) | 3.125(4) | 4 (7) |

TABLE IV
EXPERIMENTAL RESULTS OF AVERAGE NUMBER OF EQUIVALENT FEATURE SUBSETS ON FEATURE SELECTION PROBLEMS

| Dataset | HREA | MMEA-WI | MMOEA/DC | CPDEA | TriMOEA-TA&R | MO_Ring_PSO_SCD | Omni-optimizer |
|---|---|---|---|---|---|---|---|
| $FS\text{-}D_1$ | 0(4) | 2(2) | 0(4) | **5(1)** | 1(3) | 0(4) | 0(4) |
| $FS\text{-}D_2$ | 1(3) | 3(2) | 0(4) | 1(3) | 3(2) | 0(4) | **6(1)** |
| $FS\text{-}D_3$ | 0(2) | 0(2) | 0(2) | 0(2) | 0(2) | 0(2) | **2(1)** |
| $FS\text{-}D_4$ | **6(1)** | 0(3) | 0(3) | 0(3) | 0(3) | 0(3) | 1(2) |
| $FS\text{-}D_5$ | 0(2) | **1(1)** | 0(2) | 0(2) | 0(2) | 0(2) | 0(2) |
| $FS\text{-}D_6$ | **0(1)** | **0(1)** | **0(1)** | **0(1)** | **0(1)** | **0(1)** | **0(1)** |
| $FS\text{-}D_7$ | **0(1)** | **0(1)** | **0(1)** | **0(1)** | **0(1)** | **0(1)** | **0(1)** |
| $FS\text{-}D_8$ | 1(3) | **5(1)** | 1(3) | **5(1)** | 4(2) | **5(1)** | 0(4) |
| average ranking | 2.125(4) | **1.625(1)** | 2.5(6) | 1.75(2) | 2(3) | 2.25(5) | 2(3) |

TABLE V
DETAILS OF LOCATION SELECTION DATASETS

| Dataset | District | Central Place | Number of Primary Schools | Number of Middle Schools | Number of Shopping Centers | Number of Subway Stations |
|---|---|---|---|---|---|---|
| $LS\text{-}D_1$ | Tianhe District | South China Normal University | 40 | 23 | 27 | 17 |
| $LS\text{-}D_2$ | Haizhu District | Sun Yat-sen University | 50 | 30 | 14 | 12 |
| $LS\text{-}D_3$ | Yuexiu District | Tianhe City, Beijing Road | 98 | 61 | 34 | 22 |
| $LS\text{-}D_4$ | Panyu District | South China University of Technology | 7 | 1 | 4 | 4 |

TABLE VI
EXPERIMENTAL RESULTS OF AVERAGE IGDX ON LOCATION SELECTION PROBLEMS

| Dataset | HREA | MMEA-WI | MMOEA/DC | CPDEA | TriMOEA-TA&R | MO_Ring_PSO_SCD | Omni-optimizer |
|---|---|---|---|---|---|---|---|
| $LS\text{-}D_1$ | 8.02E–03(5) | 6.64E–03(2) | 6.92E–03(4) | **5.40E–03(1)** | 2.76E–02(6) | 6.84E–03(3) | 4.40E–02(7) |
| $LS\text{-}D_2$ | **5.34E–03(1)** | 6.46E–03(3) | 6.89E–03(4) | 5.58E–03(2) | 3.71E–02(6) | 7.83E–03(5) | 4.19E–02(7) |
| $LS\text{-}D_3$ | **5.90E–03(1)** | 9.58E–03(4) | 1.07E–02(5) | 6.65E–03(2) | 1.57E–02(6) | 8.55E–03(3) | 2.83E–02(7) |
| $LS\text{-}D_4$ | **6.25E–03(1)** | 7.30E–03(4) | 6.91E–03(3) | 6.55E–03(2) | 2.67E–02(7) | 9.43E–03(6) | 9.42E–03(5) |
| average ranking | 2(1) | 3.25(3) | 4(4) | **1.75(1)** | 6.25(6) | 4.25(5) | 6.5(7) |

TABLE VII
EXPERIMENTAL RESULTS OF AVERAGE IGD ON LOCATION SELECTION PROBLEMS

| Dataset | HREA | MMEA-WI | MMOEA/DC | CPDEA | TriMOEA-TA&R | MO_Ring_PSO_SCD | Omni-optimizer |
|---|---|---|---|---|---|---|---|
| $LS\text{-}D_1$ | 4.83E–03(3) | 4.36E–03(2) | 7.13E–03(5) | **4.29E–03(1)** | 1.37E–02(7) | 5.16E–03(4) | 1.15E–02(6) |
| $LS\text{-}D_2$ | **4.87E–03(1)** | 5.91E–03(3) | 7.53E–03(5) | 5.78E–03(2) | 1.82E–02(6) | 6.91E–03(4) | 1.90E–02(7) |
| $LS\text{-}D_3$ | **6.16E–03(1)** | 6.79E–03(2) | 1.02E–02(7) | 7.06E–03(3) | 8.38E–03(5) | 7.38E–03(4) | 8.67E–03(6) |
| $LS\text{-}D_4$ | **7.89E–03(1)** | 9.02E–03(4) | 8.19E–03(3) | 7.98E–03(2) | 3.27E–02(7) | 1.18E–02(6) | 1.12E–02(5) |
| average ranking | **1.5(1)** | 2.75(3) | 5(5) | 2(2) | 6.25(7) | 4.5(4) | 6(6) |

Tables III and IV show the average 1/HV and the number of equivalent feature subsets for each algorithm on each dataset, with the best results highlighted in **bold**. The numbers in parentheses next to the data indicate the ranking of each algorithm. The last row calculates the average ranking of each algorithm to assess their overall performance.

From Table III, it can be seen that regarding 1/HV, HREA performs poorly on datasets with a high number of classes ($FS\text{-}D_1$, $FS\text{-}D_2$, $FS\text{-}D_3$). MMEA-WI is the overall best algorithm for these 8 datasets, followed by MMOEA/DC. CPDEA shows poor performance on the dataset $FS\text{-}D_4$, which has many features but few samples. TriMOEA-TA&R underperforms on datasets with a high number of features (*FS-D$_3$*, *FS-D$_4$*). MO_Ring_PSO_SCD does not perform well on datasets with many classes or features (*FS-D$_1$*, *FS-D$_2$*, *FS-D$_3$*, *FS-D$_4$*), but its overall score is better than that of TriMOEA-TA&R. Omni-optimizer, as an early algorithm, is the overall worst-performing algorithm. However, it performs well on datasets with few features and few classes (*FS-D$_5$*, *FS-D$_6$*, *FS-D$_7$*, *FS-D$_8$*). By observing the number of equivalent feature subsets in Table IV, we can see that MMEA-WI finds a higher number of equivalent feature subsets across more datasets, thus achieving the top overall ranking. CPDEA also identifies a significant number of equivalent feature subsets. Omni-optimizer, being an early algorithm that emphasizes diversity, obtains a decent ranking. However, focusing on diversity may lead to reduced convergence, which explains its poor performance in terms of 1/HV. TriMOEA-TA&R outperforms MO_Ring_PSO_SCD in finding equivalent feature subsets, indicating the effectiveness of its diversity strategy. As algorithms that preserve local PFs, HREA and MMOEA/DC might not be as effective in searching global PSs as other algorithms that focus only on the global PF. Although MMOEA/DC performs well in terms of 1/HV, it only finds one equivalent feature subset across the 8 datasets. In conclusion, MMEA-WI is the most suitable algorithm for the selected feature selection problems.

### C. Experimental Results on Multimodal Multiobjective Location Selection

The details of the four datasets applied in this experiment are shown in Table V. The population size of the tested algorithms is set to 200, and the maximum number of function evaluations is set to 20000. Each algorithm is repeated 21 times on each dataset and average results are presented. The other parameters are set according to the recommendations in the corresponding references of the tested algorithms.

We use inverted generalized distance in the decision space (IGDX) and inverted generalized distance in the objective space (IGD) [33] as evaluation metrics, defined as follows:

$$\text{IGDX} = \frac{\sum_{Y \in S\_dec} \min_{X \in S} d\_dec(X,Y)}{|S\_dec|} \quad (2)$$

$$\text{IGD} = \frac{\sum_{Y \in S\_obj} \min_{X \in S} d\_obj(X,Y)}{|S\_obj|} \quad (3)$$

First, 300*300 points are uniformly sampled in the decision space. Then the non-dominated points are selected to obtain the reference points for the ideal PS and PF, denoted as ***S_dec*** and ***S_obj***, respectively. The Euclidean distance between two solutions ***X*** and ***Y*** in the decision space and objective space are represented by *d_dec* (***X, Y***) and *d_obj* (***X, Y***), respectively. IGDX and IGD represent the algorithm's performance in the decision space and objective space, respectively, and both metrics are better when smaller. To obtain a more accurate comparison of IGD, we retain the floating-point values of each objective (i.e., the distances from the house to the nearest primary school, middle school, shopping center, and subway station) and use these floating-point values to calculate IGD.

The average IGDX and IGD for each algorithm on each dataset are shown in Tables VI and VII, respectively, with the best results highlighted in **bold**. The rankings of the algorithms are indicated in parentheses next to the data. To evaluate the overall performance of each algorithm, we calculate the average ranking, which is displayed in the last row of the tables.

From Tables VI and VII, it can be observed that HREA and CPDEA perform well, achieving the best results on three and one datasets, respectively. Note that HREA's method considers local Pareto optimal solutions, which may weaken its search for global Pareto optimal solutions. This leads to HREA missing some global Pareto optimal solutions on the first dataset, resulting in a lower ranking. MMEA-W's IGDX and IGD results are second only to HREA and CPDEA. MO_Ring_PSO_SCD, with its ring topology and crowding distance strategies in both decision and objective spaces, achieves good results despite being a classic algorithm. MMOEA/DC, which employs dual clustering in the decision and objective spaces to preserve both global and local PFs, lacks sufficient search for global PF, resulting in a lower overall IGD ranking compared to MO_Ring_PSO_SCD. TriMOEA-TA&R uses two archives (i.e., convergence and diversity archives) and recombination strategies. However, its diversity archive prioritizes objective space diversity, possibly missing some Pareto optimal solutions in the decision space. Additionally, TriMOEA-TA&R's niching strategy struggles to distinguish between unequally distant Pareto optimal solutions, leading to missed solutions. Consequently, TriMOEA-TA&R performs worse than MO_Ring_PSO_SCD across the four datasets. Omni-optimizer, a classic algorithm designed for MMOPs, shows poor performance, but it lays the foundation for many subsequent algorithms. For instance, the crowding distance strategies in the decision and objective spaces of Omni-optimizer are reflected in MO_Ring_PSO_SCD. In summary, HREA and CPDEA have promising performance on location selection problems.

### IV. CONCLUSION

The proliferation of MMOPs has led to extensive research on MMOAs. Most studies evaluate algorithm performance based on benchmark function sets, particularly the CEC'2019. However, benchmark functions of MMOPs typically have low dimension sizes in both decision and objective spaces, and are weakly correlated with real-world scenarios. Therefore, this paper tests four state-of-the-art and three classic MMOAs on two types of real-world MMOPs (i.e., feature selection and location selection). For the feature selection problems, eight datasets with varying numbers of samples, features, and classes are chosen to compare the effectiveness of different algorithms. For the location selection problems, four real-world datasets based on Guangzhou, China are generated. Finally, we analyze the experimental results to discuss the features of the tested MMOAs and identify suitable algorithms for each problem. These analyses provide references for future research.